\pdfoutput=1

\documentclass[11pt]{article}
\usepackage[table]{xcolor}

\usepackage[]{acl}

\usepackage{times}
\usepackage{latexsym}
\usepackage{multirow} 
\usepackage{arydshln}

\usepackage{subfig}
\usepackage{verbatimbox}
\usepackage{lipsum} 
\usepackage{enumitem}
\usepackage{moreverb}

\usepackage[T1]{fontenc}

\usepackage[utf8]{inputenc}

\usepackage{microtype}

\usepackage{inconsolata}

\usepackage{graphicx}

\title{Step-by-Step Fact Verification System for Medical Claims \\ with Explainable Reasoning}

\author{Juraj Vladika, Ivana Hacajová, Florian Matthes \\
  Technical University of Munich, Germany \\
  School of Computation, Information and Technology \\
  Department of Computer Science \\
  \texttt{\{juraj.vladika, ivana.hacajova, matthes\}@tum.de}}

\begin{document}
\maketitle
\begin{abstract}
Fact verification (FV) aims to assess the veracity of a claim based on relevant evidence. The traditional approach for automated FV includes a three-part pipeline relying on short evidence snippets and encoder-only inference models. More recent approaches leverage the multi-turn nature of LLMs to address FV as a step-by-step problem where questions inquiring additional context are generated and answered until there is enough information to make a decision. This iterative method makes the verification process rational and explainable. While these methods have been tested for encyclopedic claims, exploration on domain-specific and realistic claims is missing. In this work, we apply an iterative FV system on three medical fact-checking datasets and evaluate it with multiple settings, including different LLMs, external web search, and structured reasoning using logic predicates. We demonstrate improvements in the final performance over traditional approaches and the high potential of step-by-step FV systems for domain-specific claims. 
\end{abstract}

\section{Introduction}
The digital age has been marked by the rise and spread of online misinformation, which has negative societal consequences, especially when related to public health \cite{vanderLinden2022MisinformationSS}. Fact verification (FV) has emerged as an automated approach for addressing the increasing rate of deceptive content promulgated online \cite{das2023state, schlichtkrull-etal-2023-intended}. On top of that, FV can help improve the factuality of generative large language models \cite{DBLP:journals/natmi/AugensteinBCCCCDFHHHJMM24} and help scientists find reliable evidence for assessing their research hypotheses  \cite{eger2025transformingsciencelargelanguage}.

The common pipeline for automated fact verification consists of document retrieval, evidence extraction, veracity prediction, and optionally justification production \cite{guo2022survey}. In such a setup, document retrieval is usually done with a method like BM25 or semantic search, evidence selected using sentence embedding models, and the final verdict predicted with an encoder-only model like DeBERTa \cite{he2021deberta}. In fact, most state-of-the-art FV systems for the popular FEVER dataset \cite{thorne-etal-2018-fever} and other recent real-world misinformation datasets rely on this pipeline  \cite{zhang-etal-2024-need, glockner-etal-2024-ambifc}.

\begin{figure}[t]
    \centering
    \includegraphics[width=0.815\linewidth]{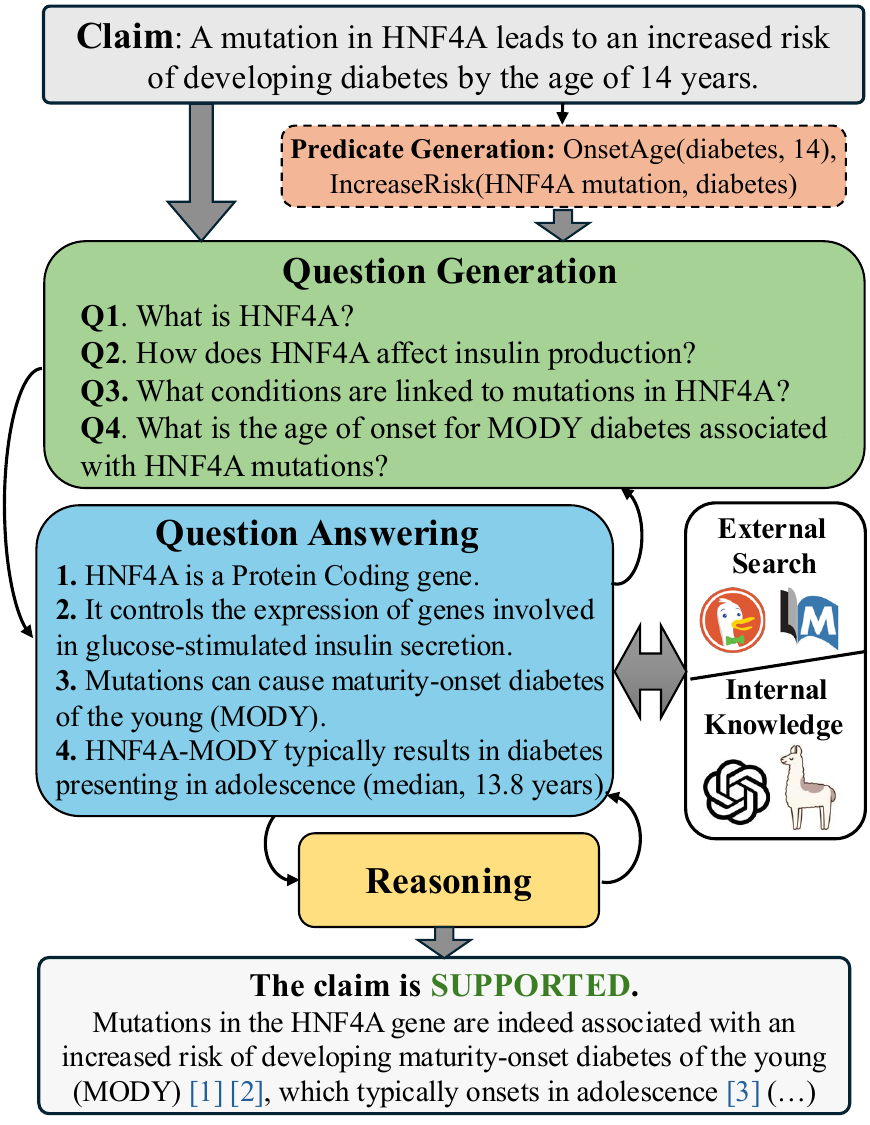}
    \caption{The step-by-step fact verification system used in our study iteratively collects additional knowledge and evidence until it can predict a veracity verdict.}
    \label{fig:system}
\end{figure}

Similarly, most previous work relies on providing pre-selected evidence to the final inference model.
A more realistic setting is \textit{open-domain}
fact verification, where evidence first has to be discovered in large knowledge bases before the system produces the verdict. Recent FV work has explored this setting, but most of them also rely on the traditional pipeline, utilizing BM25, sentence embeddings, and encoder-only inference model for producing their verdicts \cite{wadden-etal-2022-scifact, stammbach2023choice, vladika-matthes-2024-improving}.

The recent advent of large language models (LLMs) has transformed the field of NLP \cite{fan2024bibliometric}. LLMs have many properties that positively benefit the fact-verification process \cite{dmonte2025claimverificationagelarge}. First, their long context window means a lot more evidence can be provided than to encoder-only models.  Furthermore, the multi-turn nature of instruction-tuned LLMs has enabled addressing FV as a step-by-step problem where new questions inquiring for more evidence are generated in subsequent iterations before there is enough information to produce a verdict on a claim's veracity \cite{dhuliawala-etal-2024-chain}.  
This also makes the verification process interpretable since the reasoning steps can be traced through the question-answer pairs, thus justifying the verdict \cite{eldifrawi-etal-2024-automated}. 

These step-by-step LLM systems for FV have been shown to work well on complex, multi-hop claims found in datasets like HOVER \cite{jiang-etal-2020-hover}. Intuitively, complex synthetic claims from these datasets, like "\textit{Yao Ming's wife's alma mater is in Texas}", have to be broken down into sub-units to be verified effectively. Nevertheless, we posit that more realistic but simple claims such as "\textit{Honey can cure a common cold}" also necessitate generating follow-up questions and collecting deeper knowledge before producing a verdict. To the best of our knowledge, no research has been conducted to test how well can these step-by-step FV systems perform on domain-specific claims. 

To bridge this research gap, in this study, we develop a step-by-step LLM system, shown in Figure \ref{fig:system}, and apply it on three medical fact-checking datasets. We contrast the results to the previous work on open-domain scientific fact verification based on a traditional system, showcasing significant improvements in the final predictive performance of the system. We outline additional findings regarding the influence of the base LLM, evidence source, and reasoning with predicate logic on the final verification performance, highlighting the great potential of these systems for diverse claims.

We make our data and code available in a public GitHub repository.\footnote{\url{https://github.com/jvladika/StepByStepFV}}

\section{Related Work}
There have been many synthetic FV datasets constructed from Wikipedia, such as FEVER \cite{thorne-etal-2018-fever}. While FEVER focuses on simple claims, datasets like 
HOVER \cite{jiang-etal-2020-hover} and 
FEVEROUS \cite{aly-etal-2021-fact} introduced complex claims requiring multi-hop reasoning. Apart from synthetic datasets, there are also datasets focusing on more realistic claims and real-world misinformation \cite{schlichtkrull2023averitec, glockner-etal-2024-ambifc}. Increasingly popular are also domain-specific datasets focusing on scientific fact-checking \cite{vladika-matthes-2023-scientific}, especially for the domains of medicine \cite{saakyan-etal-2021-covid, sarrouti2021evidence}, climate \cite{diggelmann2020climatefever}, and computer science \cite{lu-etal-2023-scitab}. 

Most FV approaches follow the traditional three-part pipeline \cite{bekoulis2021review}. In recent years, approaches incorporating LLMs and iterative reasoning into the process have achieved great performance on multi-hop FV. This includes FV through varifocal questions \cite{ousidhoum-etal-2022-varifocal} or \textit{wh}-questions to aid verification \cite{rani-etal-2023-factify}, step-by-step prompting \cite{zhang-gao-2023-towards}, and program-guided reasoning \cite{pan-etal-2023-fact}.

Most studies with iterative FV systems focus on multi-hop encyclopedic claims. To the best of our knowledge, our study is among the first to explore the step-by-step FV systems for real-world claims rooted in scientific and medical knowledge. 

\section{Foundations}
In this section, we describe in more detail the two FV approaches: the conventional three-part pipeline, serving as a baseline, and the step-by-step LLM-based system, which we mainly use.

\subsection{Three-Part Pipeline for Fact Verification}
The traditional three-part pipeline consists of: (1) document retrieval; (2) evidence extraction; (3) verdict prediction. It was used in the study by \citet{vladika-matthes-2024-comparing}, whose results we use as the baseline. Since it is an open-domain FV system, evidence documents have to be retrieved first. For that, step (1) was modeled with semantic search (similarity of query and corpus embeddings) over a large document corpus (PubMed and Wikipedia). In another experiment, evidence was sought with Google search. After selecting the top documents, step (2) again used a sentence embedding model to compare the claim to passages from the documents, selecting the most relevant evidence snippets. Finally, step (3) is modeled as the task of Natural Language Inference (NLI), where the goal is to predict the logical entailment relation between the claim and evidence, i.e., whether the claim is supported by evidence (entailment), refuted by evidence (contradiction), or there is not enough information (neutral). The model was DeBERTa-v3 fine-tuned on various NLI datasets from \citet{laurer2024less}.

\subsection{Step-by-Step LLM System}
The recent LLM advancements have brought a lot of features that can enhance the FV process. With their generative capabilities and multi-turn nature, LLMs can generate follow-up questions that aim to collect deeper background evidence related to claims. 
They are able to produce verdicts for claims over multiple pieces of evidence with mechanisms like chain-of-thought reasoning \cite{ling2023deductive}.

The system we develop in this work is mainly inspired by QACheck \cite{pan-etal-2023-qacheck} and its FV components. We expand that system by introducing novel prompts, additional chain-of-thought reasoning, amplify evidence retrieval with an online search engine, and experiment with structured reasoning in the form of logic predicates. The idea of this system is, given the claim \textit{c} being verified, to generate up to five follow-up questions $q_1, ..., q_5$, which try to gather more evidence related to the claim. This is generated using a base LLM $M_q$ and a prompt. Afterward, evidence for each question $q$ is retrieved from the source $s$ (web search or internal knowledge) using the method $R(q,s)$. This collected evidence is summarized with model $M_s$ and together with original $c$ posed to a reasoning model $M_r$. This reasoning module determines whether it should continue generating new questions or if there is enough evidence. If there is enough, it predicts a final verdict label $v$, one of \textsc{Supported} or \textsc{Refuted}, and generates an explanation $e$.

On top of the described approach, we also experiment with a setting incorporating \textit{predicate logic} into the process. Given the claim $c$, a predicate is generated by an LLM in the form of \textit{verb(subject, object)}, such as \textit{Treats(aspirin, headache)}, and used to generate better questions $q_i$ and verdict $v$. Inspired by FOLK \cite{wang-shu-2023-explainable}, the idea behind this is that the structured nature of predicates can help in finding more accurate evidence and introduce structured reasoning for the final verdict prediction \cite{strong-etal-2024-zero}. 

\section{Experiments and Setup}

In the experiments, our main research question is \textbf{RQ:} \textit{Does the iterative LLM approach outperform the traditional three-part pipeline for domain-specific fact verification?} On top of that, we test three further aspects of the system: (a) knowledge source, (b) structured reasoning, and (c) base LLM. 

The knowledge sources include: internal knowledge of the LLM and the online search of the whole web. Our search engine of choice is DuckDuckGo, an open-source tool focused on privacy. We use it through a dedicated Python library.\footnote{\url{https://pypi.org/project/duckduckgo-search/}} This search engine provided a smooth search experience with no interruptions, and we deemed the quality of the retrieved results similar to the more popular Google or Bing for our use case. We take the provided \textit{snippets} from the first 5 results and give them as input evidence to the reasoner LLM. The structured reasoning in (b) refers to using logic predicates, as described in the previous section. All the experiments in (a) and (b) were done using \textit{GPT-4o-mini-2024-07-18} as the base LLM, the model from OpenAI with good reasoning capabilities \cite{openai2024gpt4technicalreport}.

In experiment round (c), we additionally test normal reasoning with internal knowledge and online search using Mixtral 8x7B \cite{jiang2024mixtralexperts}, a highly performing open-weights model based on a mixture-of-experts architecture, and LLaMa 3.1 (70B) \cite{dubey2024llama3herdmodels}, a recent advanced open-weights model from Meta. We use GPT through the OpenAI API and the two other models through the Together AI API,\footnote{\url{https://www.together.ai}} setting temperature to 0 for best reproducibility and maximum tokens to 512. We use these LLMs for all parts of the fact verification process, i.e. for all steps $M_q, M_s, M_r$ as described in the previous section. All the used prompts are in the Appendix. All experiments were run on one Nvidia V100 GPU with 16 GB VRAM.

\begin{table*}[htpb]
\footnotesize
	\centering
	\rowcolors{2}{blue!10}{blue!2}
	\resizebox{\textwidth}{!}{%
		
		\begin{tabular}{c|c|ccc|ccc|ccc}
			
	\rowcolor{blue!25}
	\textbf{verification} & \textbf{evidence} & \multicolumn{3}{c}{\textbf{HealthFC}} & \multicolumn{3}{c}{\textbf{CoVERT}} & 
	\multicolumn{3}{c}{\textbf{SciFact}}  \\
	
	\rowcolor{blue!25}
	\textbf{system} &  \textbf{source}  & \textbf{P}       & \textbf{R}       & \textbf{F1}  & \textbf{P}       & \textbf{R}       & \textbf{F1}       & \textbf{P}       & \textbf{R}       & \textbf{F1}     \\
	\hline
 \textbf{Three-part pipeline} & PubMed  & $62.6$ & $84.6$ & $72.0$  & $75.6$ & $76.8$ & $76.2$ & $73.7$ & $80.0$ & $76.8$ \\ 
  (with semantic search & Wikipedia  & $65.2$ & $92.6$ & $76.5$ & $78.5$ & $86.8$ & $82.5$ &  $68.8$ & $83.6$ & $75.4$ \\ 
  and DeBERTa) & whole web  & $62.3$ & $92.6$ & $74.5$  & $76.4$ & $68.7$ & $72.3$  & $75.5$ & $91.5$ & $82.7$  \\ \hline \hline
	\textbf{GPT 4o-mini system} &  whole web  & 71.4 & 90.1 & 79.6  & 88.7 & 83.3 & \textbf{85.9}  & 87.7 & 87.5 & \textbf{87.6}  \\ 
	 &  internal  & 72.3 & 91.6 & \underline{80.8}  & 87.4 & 80.8 & 84.0  & 83.5 & 82.5 & 83.0  \\ \hline
	\textbf{GPT 4o-mini system} & whole web   & 74.9 & 88.6 & 81.2  & 90.1 & 68.7 & 77.9  & 88.2 & 82.2 & \underline{85.1}  \\ 
	(with predicates) &  internal  & 73.7 & 91.6 & \textbf{81.7}  & 89.1 & 70.2 & 78.5  & 84.9 & 77.9 & 81.2  \\ \hline
	\textbf{Mixtral 8x7B system} &  whole web  & 68.2 & 78.7 & 73.1  & 79.8 & 81.8 & 80.8  & 82.0 & 86.2 & 84.1  \\
	 & internal  & 68.5 & 74.3 & 71.3  & 86.9 & 77.3 & 81.8 & 80.9 & 83.3 & 82.1  \\ \hline
	\textbf{LLaMa 3.1 (70B) system} &  whole web  & 74.3 & 88.6 & \underline{80.8}  & 79.1 & 89.9 & \underline{84.2}  & 86.1 & 82.7 & 84.3  \\ 
      & internal  & 64.7 & 86.1 & 73.9  & 74.3 & 81.8 & 77.9 & 80.0 & 87.5 & 83.6  \\ \hline

		\end{tabular}
	}
	
	\caption{\label{tab:results} The results of the study. The first three rows come from a related study using the three-part pipeline. The further rows are from this study, using a consistent system with varying base LLM, structured reasoning type, and evidence source. The best F1 score for each dataset is in \textbf{bold}, while the second best is \underline{underlined}.}
	
\end{table*}

\subsection{Datasets and Evaluation}
We choose three English datasets of biomedical and healthcare claims, designed for different purposes:

 \textsc{SciFact} \cite{wadden-etal-2020-fact} is a dataset with expert-written biomedical claims originating from citation sentences found in medical paper abstracts. The subset we use contains 693 claims, of which 456 are supported, and 237 are refuted.
 
\textsc{HealthFC} \cite{vladika-etal-2024-healthfc} is a dataset of claims concerning everyday health and spanning various topics like nutrition, the immune system, and mental health. The claims originate from user inquiries and they were checked by a team of medical experts. The subset we use contains 327 claims, of which 202 are supported, and 125 are refuted.

 \textsc{CoVert} \cite{mohr-etal-2022-covert} is a dataset of health-related claims, which are all causative in nature (such as "\textit{vaccines cause side effects}"). All the claims originate from Twitter, which brings an additional challenge of informal language and provides a real-world scenario of misinformation checking. The subset we use contains 264 claims, of which 198 are supported, and 66 are refuted.

We find these three datasets to be well suited for our study because they are representative of three different applications of fact verification: helping researchers in their work (\textsc{SciFact}), verifying everyday user questions (\textsc{HealthFC}), and misinformation detection on social media (\textsc{CoVERT}). 

We take claims from these datasets and use them as input to our system. To evaluate if the prediction is correct, we use the original veracity gold label. We do not give the system any original gold evidence documents from the datasets, as we are studying an open-domain setting. In essence, we evaluate the performance of the whole system by looking at its final classification performance as a "proxy" and observing how it changes when varying different parameters \cite{chen-etal-2024-complex}.  While an important class in datasets is \textit{not enough information} (NEI), we simplify the problem to only the \textit{supported} and \textit{refuted} classes and leave NEI for future work. Therefore, we use binary precision, recall, and F1 score as the evaluation metrics.

\begin{table*}[htpb]
\small
\centering
\begin{tabular}{p{13.5mm}p{40mm}p{42.5mm}p{47mm}}
\hline

\rowcolor{blue!25}
\small \textbf{ }    & \hfil \small  \textbf{GPT 4o-mini}    & \hfil \small \textbf{ LLaMa 3.1 }  &  \hfil \small \textbf{ Mixtral 8x7B }     \\ \hline

\rowcolor{blue!5}
\raggedright \textbf{Claim} \newline (Dataset)     &  EMDR works similarly to exposure therapy, another well-studied form of psychotherapy, for the symptoms of post-traumatic stress disorder. \small (\textsc{\textbf{HealthFC}})
&  The relationship between a breast cancer patient's capacity to metabolize tamoxifen and treatment outcome is independent of the patient's genetic make-up. \small (\textsc{\textbf{SciFact}})
&  Severe Covid is a systemic disease not just confined to resp system... causes hyper inflammation,clotting, encephalopathy etc..we need all hands on deck from generalists to specialists, nurses, ward attendants! \small (\textsc{\textbf{CoVert}}) \\ \hline

\rowcolor{blue!5}                     
\raggedright \textbf{Generated Questions}        & 
\scriptsize
\textbf{Q1:} How does EMDR therapy work? \newline
\textbf{Q2:} What are the key components of exposure therapy for PTSD? \newline
\textbf{Q3:} What are the similarities between EMDR and exposure therapy for PTSD? \newline
\textbf{Q4:} What are the differences between EMDR and exposure therapy for PTSD? \newline
\textbf{Q5:} What are the outcomes of EMDR compared to exposure therapy for PTSD?
& 
\scriptsize

\textbf{Q1:} What is tamoxifen used for in breast cancer treatment? \newline 
\textbf{Q2:} Is the metabolism of tamoxifen influenced by genetic factors? \newline 
\textbf{Q3:} What does research suggest about tamoxifen metabolism in breast cancer patients with different genotypes?  \newline 
\textbf{Q4:} Does the genetic variation in tamoxifen metabolism affect the treatment outcome in breast cancer patients?  
&

\scriptsize
\textbf{Q1:} What are some of the problems beyond respiratory issues that severe Covid causes? \newline
\textbf{Q2:} What types of medical professionals are needed to treat severe Covid? \newline
\textbf{Q3:} What medical equipment is used for patients with severe Covid in intensive care? \newline
\textbf{Q4:} How is hyperinflammation treated in Long Covid patients? \newline
\textbf{Q5:} How is anticoagulation managed in Long Covid patients to prevent clotting?
\\ \hline

\end{tabular}
\caption{\label{tab:top3} Examples of three claims (all \textcolor{teal}{supported}) from the three datasets used in the study, with generated verification questions from the three different LLMs. GPT generates the most general questions with wider coverage, while LLaMa and Mixtral generate more specific and in-depth questions. }
\end{table*}

\section{Results and Discussion}

The first three rows of Table \ref{tab:results} show the results of the traditional three-part pipeline (described in Section 3.1) taken from the related study by \citet{vladika-matthes-2024-comparing}. It compared the performance over three knowledge sources: PubMed, Wikipedia, and online search.
The results in further rows are from the experiments done in this study. 

\paragraph{Improvement.} As seen in Table \ref{tab:results}, the step-by-step verification systems considerably improved the final F1 performance on all three datasets, especially precision values. The first GPT system improved the F1 performance by +4.3 on HealthFC, +3.4 on CoVERT, and +4.9 on SciFact, which is a major improvement when compared to the traditional pipeline using single-turn verification. This answers our main research question.

\paragraph{Internal vs.~External Knowledge.} 
Utilizing web search improved the performance in all cases for SciFact, showing that this dataset worked better when grounded to biomedical studies found online. For the other two datasets, which contain common health claims, there were instances where internal knowledge of LLMs even outperformed the web search. This is a very noteworthy finding, demonstrating how LLMs already encode a lot of internal medical knowledge that can be useful in knowledge-rich tasks, as observed by \citet{Singhal2023-vc} and \citet{vladika-etal-2024-medreqal}. Similarly, \citet{frisoni-etal-2024-generate} showed how using LLM-generated evidence passages can improve medical QA performance more than retrieved passages.

\paragraph{Predicate Logic.} The next experiment incorporated first-order-logic predicates into the FV process. In the GPT system, this resulted in the best overall performance for HealthFC, ending at 81.7 F1 (+5.2 improvement to baseline, +1 to without predicates). This is because predicates, like \textit{Outcomes(Tamoxifen, Breast Cancer)}, led to more precise and targeted evidence, as indicated by the increase in precision scores.
On the other hand, while precision also increased for the other two datasets, it led to large drops in recall, resulting with a lower F1. This was especially seen with informal language in CoVERT claims, where produced predicates included underspecified instances like \textit{Has(Person, Covid)}, which only degraded the evidence retrieval process. Therefore, predicates are better suited for clearly written queries and for complex claims.

\paragraph{Choice of LLMs.} Comparative analysis of different LLMs was the last round of experiments. Overall, GPT-4o-mini came out on top as the best LLM for the task. Table \ref{tab:top3} shows an example of generated questions for all three LLMs for different claims. It is evident that GPT gives the most general and simplest questions, whereas LLaMa and Mixtral provide more specific and detailed questions. The specific questions can be a strength but also complicate the evidence retrieval process with noisy retrieved passages. GPT was the best at following the style of few-shot example questions.  Also, Mixtral produces the most questions on average per claim, followed by GPT, and then LLaMa.  Finally, we observed the reasoning capabilities of models to be on a similar level, showing the final performance is often dependent on the quality of question generation and answering.

\paragraph{Qualitative Analysis.} As evident in Table \ref{tab:top3}, a lot of generated questions were asking for definitions of the diseases, symptoms, drugs, and other terms found in claims. Once such complex terms were described, the FV process was well-equipped to continue with the verification. This explains \textit{why} the step-by-step systems worked so well for medical claims, similarly to multi-hop claims in previous studies -- they inherently contain complex concepts and relations that shall be clarified first before making the final decision. 

A common reason for errors in the system was the generated questions going too in-depth about a certain point with its follow-up questions and not collecting wider evidence about other parts of the claim. Moreover, another issue were \textit{knowledge conflicts} -- when the LLM would predict an incorrect label even when shown evidence to the contrary because of its encoded internal knowledge. 

Future work could expand the system to leverage structured knowledge sources like knowledge graphs \cite{kim-etal-2023-factkg} or use methods like formal proof generation \cite{strong-etal-2024-zero}. The final step of the system focusing on explanation generation should ideally include different user perspectives in the process \cite{warren2025show}.


\section{Conclusion}
In this study, we develop a step-by-step system for fact verification based on iterative question generation and explainable reasoning. We apply the system on three medical fact-checking datasets and test different settings. We show that by utilizing LLMs, this system can create follow-up questions on complex concepts and relations from the claims in order to gather background evidence, reason over newly discovered evidence, and finally lead to predictions that achieve higher results when compared to traditional pipelines. We hope that our study encourages more exploration of advanced systems for domain-specific fact verification.

\section*{Limitations}
Since all modules of the step-by-step verification system rely on using LLMs, they come with their own set of challenges and limitations. The generated follow-up questions are not always perfect or precise, the generated evidence snippets can be off point, and the reasoning over long chains of evidence can, of course, lead to logical errors and mistakes. We observed certain instances where even though all the evidence was pointing towards one of the verdicts (\textit{refuted}), the system would still mistakenly output the other one (\textit{supported}).

Another limitation comes from the high complexity of the system and reliance on calls to external APIs, including LLM APIs and search engine APIs. This inevitably led to some challenges in terms of slower processing speed of this system when compared to traditional approaches that use an out-of-the-box NLI model like DeBERTa. Still, we were forced to rely on API calls for LLMs due to hardware resource limitations, but models like Mixtral and LLaMa showed decent performance and are open-weights, so they can be downloaded and run locally to speed up the performance.

Lastly, for easier evaluation we disregard claims annotated with \textit{Not Enough Information} due to different definitions of this label across different datasets (e.g., the definition from SciFact does not serve the open-domain setting well). This is an important label in fact verification, since not all claims can be conclusively assessed for their veracity. This is especially important in the scientific domain considering the constantly evolving nature of scientific knowledge, and sometimes conflicting evidence from different research studies. Future work should find a way to effectively include this label into model predictions. 

\section*{Ethics Statement}
Our dataset and experiments deal with the highly sensitive domain of healthcare and biomedical NLP. While we observed good scores when verifying health-related question using responses directly generated by language models, this is not a recommended way of using them by end users or patients. Responses can still contain hallucinations or misleading medical advice that should always be manually verified within reliable sources. Similarly, experiments using online search results did not go through any manual quality filtering, which means not all of them will be trustworthy or approved by experts. One should always consult with medical professionals when dealing with health-related questions and advice. 

\section*{Acknowledgements}
This research has been supported by the German Federal Ministry of Education and Research (BMBF) grant 01IS17049 Software Campus 2.0 (TU München). We would like to thank the anonymous reviewers for their valuable feedback.

\bibliography{custom}

\appendix

\section{Appendix}
\label{sec:appendix}

In the appendix, we provide the prompts used for the systems (Figures 2--7).

\begin{verbbox}[\mbox{}]
Claim = Superdrag and Collective Soul are both rock bands.
To validate the above claim, the first simple question we need to ask is:
Question = Is Superdrag a rock band?

Claim = Jimmy Garcia lost by unanimous decision to a professional boxer that 
challenged for the WBO lightweight title in 1995. 
To validate the above claim, the first simple question we need to ask is: 
Question = Who is the professional boxer that challenged for the WBO 
lightweight title in 1995? 
\end{verbbox}
\begin{figure*}[htbp]
  \centering
  \fbox{\theverbbox}
  \caption{Two out of ten few-shot examples used in the prompt for generating the first verification question.}
  \label{fig:qg-module-first}
\end{figure*}

\begin{verbbox}[\mbox{}]
Claim = Superdrag and Collective Soul are both rock bands.
To validate the above claim, we need to ask the following simple questions 
sequentially: 
Question 1 = Is Superdrag a rock band?
Answer 1 = Yes
Question 2 = Is Collective Soul a rock band?

Claim = Jimmy Garcia lost by unanimous decision to a professional boxer that 
challenged for the WBO lightweight title in 1995. 
To validate the above claim, we need to ask the following simple questions 
sequentially: 
Question 1 = Who is the professional boxer that challenged for the 
WBO lightweight title in 1995? 
Answer 1 = Orzubek Nazarov
Question 2 = Did Jimmy Garcia lose by unanimous decision to Orzubek Nazarov?
\end{verbbox}
\begin{figure*}[htbp]
  \centering
  \fbox{\theverbbox}
  \caption{Two out of ten few-shot examples used in the prompt for generating the follow-up questions (after the first one had been generated).}
  \label{fig:qg-module-followup}
\end{figure*}

\begin{verbbox}[\mbox{}]
Claim = Superdrag and Collective Soul are both rock bands.
To validate the above claim, we have asked the following questions: 
Question 1 =to explainAnswer 1 = Yes
Can we know whether the claim is true or false now?
Prediction = No, we cannot know. 

Claim = Superdrag and Collective Soul are both rock bands.
To validate the above claim, we have asked the following questions: 
Question 1 = Is Superdrag a rock band?
Answer 1 = Yes
Question 2 = Is Collective Soul a rock band?
Answer 2 = Yes
Can we know whether the claim is true or false now?
Prediction = Yes, we can know.
\end{verbbox}
\begin{figure*}[htbp]
  \centering
  \fbox{\theverbbox}
  \caption{Two out of ten few-shot examples for the verifier module. In this step, the LLM decides if there is enough evidence to make the final veracity prediction or if question generation shall continue.}
  \label{fig:verifier}
\end{figure*}

\begin{verbbox}[\mbox{}]
Claim: Superdrag and Collective Soul are both rock bands.

To validate the above claim, we need to ask the first question with predicate: 
Question: 
Is Superdrag a rock band?
Predicate:
Genre(Superdrag, rock) ::: Verify Superdrag is a rock band

Claim : Jimmy Garcia lost by unanimous decision to a professional boxer that 
challenged for the WBO lightweight title in 1995. 

To validate the above claim, we need to ask the first question with predicate: 
Question: 
Who is the professional boxer that challenged for the WBO lightweight title 
in 1995? 
Predicate:
Challenged(player, WBO lightweight title in 1995) ::: Verify name of the 
professional boxer that challenged for the WBO lightweight title in 1995.
\end{verbbox}
\begin{figure*}[htbp]
  \centering
  \fbox{\theverbbox}
  \caption{Two out of ten few-shot examples for question generation in the predicate pipeline. Each generated question is accompanied by a predicate defining the question and a simple instruction on what to verify.}
  \label{fig:predicate-qg}
\end{figure*}

\begin{verbbox}[\mbox{}]
Claim: Superdrag and Collective Soul are both rock bands.

Question 1: 
Is Superdrag a rock band?
Predicate 1: 
Genre(Superdrag, rock) ::: Verify Superdrag is a rock band
Answer 1: 
Yes

To validate the above claim, we need to ask the follow-up question with predicate: 
Follow-up Question: 
Is Collective Soul a rock band?
Predicate:
Genre(Collective Soul, rock) ::: Verify Collective Soul is a rock band
\end{verbbox}
\begin{figure*}[ht]
  \centering
  \fbox{\theverbbox}
  \caption{One out of then few-shot examples of follow-up question generation for the predicate system. The already gathered evidence and predicates from previous questions are given.}
  \label{fig:predicate-qgfu}
\end{figure*}

\begin{verbbox}[\mbox{}]
Question: 
Is it true that The writer of the song Girl Talk and Park So-yeon have both 
been members of a girl group.?
Context:
Write(the writer, the song Girl Talk) ::: Verify that the writer of the song 
Girl Talk
Member(Park So-yeon, a girl group) ::: Verify that Park So-yeon is a member 
of a girl group
Member(the writer, a girl group) ::: Verify that the writer of the song Girl
Talk is a member of a gril group

Who is the writer of the song Girl Talk? Tionne Watkins is the writer of the 
song Girl Talk.
Is Park So-yeon a member of a girl group? Park Soyeon is a South Korean singer. 
She is a former member of the kids girl group I& Girls.
Is the writer of the song Girl Talk a member of a girl group? Watkins rose to 
fame in the early 1990s as a member of the girl-group TLC
Prediction:
Write(Tionne Watkins, the song Girl Talk) is True because Tionne Watkins is the
writer of the song Girl Talk.
Member(Park So-yeon, a girl group) is True because Park Soyeon is a South Korean 
singer. She is a former member of the kids girl group I& Girls.
Member(Tionne Watkins, a girl group) is True because Watkins rose to fame in the 
early 1990s as a member of the girl-group TLC
Write(Tionne Watkins, the song Girl Talk) && Member(Park So-yeon, a girl 
group) && Member(Tionne Watkins, a girl group) is True.
The claim is [SUPPORTED].
Explanation:
Tionne Watkins, a member of the girl group TLC in the 1990s, is the writer of 
the song "Girl Talk." 
Park Soyeon, a South Korean singer, was formerly part of the girl group I& Girls. 
Therefore, both Watkins and Park Soyeon have been members of girl groups in 
their respective careers.
\end{verbbox}
\begin{figure*}[htbp]
  \centering
  \fbox{\theverbbox}
  \caption{One example used in the prompt for the reasoning module using predicates.}
  \label{fig:predicate-reas}
\end{figure*}

\end{document}